\documentclass[11pt]{article}
\usepackage[margin=1in]{geometry}
\usepackage{amsmath,amssymb}
\usepackage{booktabs}
\usepackage{natbib}
\usepackage{hyperref}
\usepackage{graphicx}
\usepackage{placeins}
\usepackage{xurl}

\hypersetup{colorlinks=true,allcolors=black}

\title{Reverse Item Response Theory for Sparsity-Robust Ranking\\in Fragmented Cancer Drug-Response Matrices}
\author{Jung Min Kang\thanks{Email: \texttt{gangjeongmin23@gmail.com}}\\
Independent Researcher\\
Seoul, South Korea}
\date{May 2026}

\begin{document}
\maketitle

\begin{abstract}
We introduce reverse Item Response Theory (IRT) to pharmacogenomic drug-response analysis by treating cancer types as latent ``subjects'' with resistance ability and drugs as ``items'' with evasion difficulty. Applied to 242,036 drug sensitivity measurements from the Genomics of Drug Sensitivity in Cancer (GDSC2) database, the model estimates cancer-type-level in-vitro resistance and drug-level broad activity on a shared latent scale. Validation across four missingness regimes demonstrates that reverse IRT better recovers the full-data latent ranking than simple averaging, with advantages of $\Delta\rho = +0.089$ to $+0.095$ at 60\% missingness under MCAR, cancer-biased, and drug-biased sparsity. Held-out prediction confirms IRT achieves the best Brier score among five evaluated methods. Bootstrap confidence intervals show 19 of 28 cancer types have stable resistant/sensitive classifications. Cross-platform PRISM replication shows 82\% directional agreement but weak rank-order correlation ($\rho = 0.25$), indicating the contribution is methodological robustness under fragmented evaluation, not a universal clinical resistance leaderboard.
\end{abstract}

\section{Introduction}

Large-scale pharmacogenomic screening efforts, including the Genomics of Drug Sensitivity in Cancer \citep[GDSC;][]{yang2012gdsc,iorio2016landscape} and the PRISM Repurposing dataset \citep{corsello2020}, have generated comprehensive drug-response matrices spanning hundreds of drugs and thousands of cancer cell lines. Existing approaches include ANOVA-based biomarker discovery \citep{garnett2012}, machine learning prediction \citep{costello2014}, deep learning \citep{liu2020deepcdr}, and recent work inferring general principles of drug sensitivity with experimental validation \citep{carli2025}.

These approaches model each drug--cell-line pair independently or predict sensitivity from genomic features, without jointly estimating a cancer type's global resistance and a drug's global activity on a common measurement scale.

Item Response Theory \citep[IRT;][]{lord1968,baker2004} provides this capability. IRT jointly estimates subject ability and item difficulty on a shared latent scale. \citet{kang2026efsl} demonstrated that IRT-based ranking outperforms simple averaging under sparse evaluation in AI benchmarking. \citet{rodriguez2021} applied IRT to NLP evaluation, and \citet{polo2024} used IRT for efficient LLM benchmarking.

We propose a conceptual inversion: cancer types become ``subjects'' with latent resistance $\theta_j$, drugs become ``items'' with evasion difficulty $b_i$:
\begin{equation}
P(\text{sensitive}) = \sigma(b_i - \theta_j)
\label{eq:irt}
\end{equation}
The primary contribution is not the specific rankings---which are platform-dependent---but demonstrating that reverse IRT provides sparsity-robust ranking recovery in fragmented drug-response matrices. This is relevant because real-world therapeutic evidence matrices are sparse: drugs, indications, and trial populations are unevenly evaluated.

\section{Data}

\textbf{GDSC2.} Release 8.5 (October 2023), Wellcome Sanger Institute. Raw: 242,036 drug--cell-line measurements (969 cell lines, 286 drugs, 32 TCGA cancer types). After removing unclassified types (196,345 remaining) and excluding CLL (9 drugs, insufficient coverage), we aggregate to a $28 \times 286$ matrix with 7,821 cells (97.7\% coverage). Binarization: sensitive if fitted $\text{LN\_IC50} < 3.297$ (global median, ${\approx}27~\mu$M). A global threshold avoids drug-specific normalization; drug parameters reflect apparent broad in-vitro activity rather than absolute pharmacological potency.

\textbf{PRISM.} Secondary dose-response dataset \citep{corsello2020}: 701,004 IC50 entries, 1,448 compounds, 499 cell lines. Cell lines mapped via DepMap lineage metadata, yielding 17 overlapping cancer types.

\section{Methods}

\textbf{Reverse IRT.} 1PL model (Eq.~\ref{eq:irt}) with analytical gradients (finite-difference error $< 0.02$), L-BFGS-B optimization, Gaussian priors ($\sigma^2 = 4$). Parameters: $28 + 286 = 314$ for 7,821 observations.

\textbf{Sparsity test.} Remove 20--60\% of cells under four regimes: MCAR, cancer-biased (harder cancers lose more), drug-biased (weaker drugs lose more), pathway-block (entire pathways removed). Evaluate Spearman $\rho$ vs.\ full-data IRT ranking (15 seeds).

\textbf{Held-out prediction.} 20\% cells held out, 10-fold CV. Brier score against cancer-only averaging, drug-only averaging, two-way additive ($\text{global} + \text{cancer effect} + \text{drug effect}$), logistic fixed effects, and reverse IRT.

\textbf{Bootstrap CIs.} 200 drug-panel (column) resamples for 95\% CIs on $\theta$.

\textbf{PRISM replication.} Independent reverse IRT on PRISM; Spearman $\rho$ with GDSC2 ranking.

\section{Results}

\subsection{Sparsity Robustness}

\begin{figure}[ht]
\centering
\includegraphics[width=0.85\linewidth]{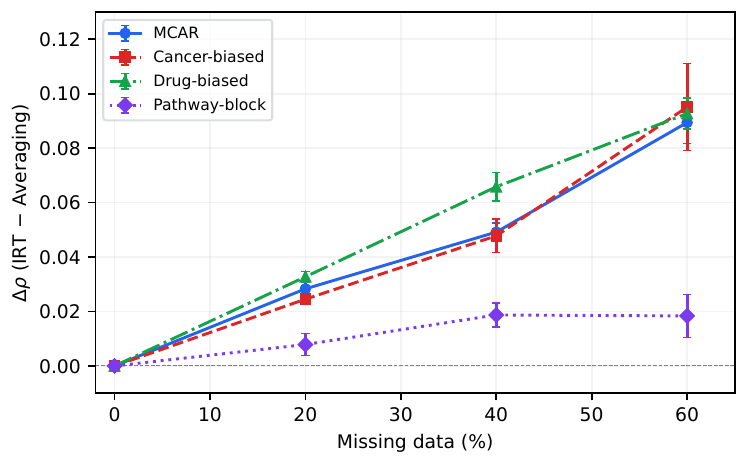}
\caption{Ranking recovery advantage ($\Delta\rho$) of reverse IRT over averaging under four missingness regimes. Error bars show $\pm 1$ SD across 15 seeds. IRT outperforms averaging in all conditions.}
\label{fig:sparsity}
\end{figure}

Reverse IRT outperforms averaging across all regimes (Table~\ref{tab:sparsity}, Figure~\ref{fig:sparsity}). The advantage generally increases with missingness, except under pathway-block missingness where it remains positive but modest.

\begin{table}[ht]
\centering
\caption{$\Delta\rho$ = IRT $-$ averaging for recovery of the full-data IRT ranking under induced sparsity (mean $\pm$ SD, 15 seeds).}
\label{tab:sparsity}
\small
\begin{tabular}{lccc}
\toprule
Regime & 20\% & 40\% & 60\% \\
\midrule
MCAR & $+0.028 \pm 0.003$ & $+0.049 \pm 0.003$ & $+0.089 \pm 0.008$ \\
Cancer-biased & $+0.025 \pm 0.002$ & $+0.048 \pm 0.006$ & $+0.095 \pm 0.016$ \\
Drug-biased & $+0.033 \pm 0.002$ & $+0.066 \pm 0.005$ & $+0.093 \pm 0.006$ \\
Pathway-block & $+0.008 \pm 0.004$ & $+0.019 \pm 0.004$ & $+0.018 \pm 0.008$ \\
\bottomrule
\end{tabular}
\end{table}

\subsection{Held-Out Prediction}

\begin{figure}[ht]
\centering
\includegraphics[width=0.8\linewidth]{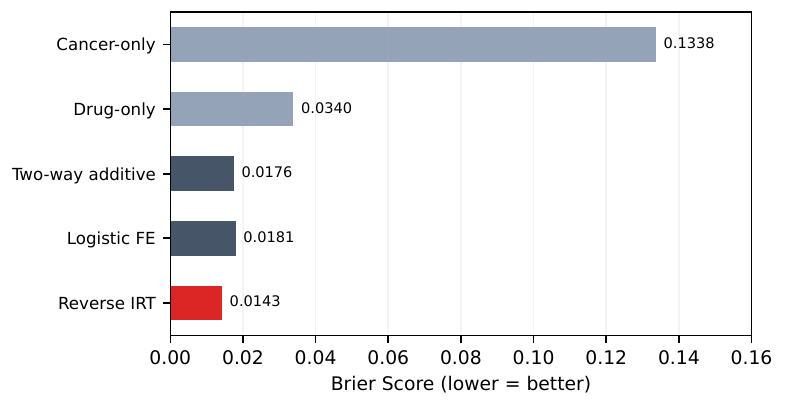}
\caption{Held-out Brier score among five evaluated methods (10-fold CV). Reverse IRT outperforms all methods including the fairer two-way additive and logistic fixed-effect baselines.}
\label{fig:heldout}
\end{figure}

IRT achieves the best Brier score among all five evaluated methods (Table~\ref{tab:brier}, Figure~\ref{fig:heldout}).

\begin{table}[ht]
\centering
\caption{Mean Brier score across 10 held-out folds (lower = better).}
\label{tab:brier}
\small
\begin{tabular}{lcc}
\toprule
Method & Brier & $\Delta$ vs IRT \\
\midrule
Cancer-only averaging & 0.1338 & $+0.1195$ \\
Drug-only averaging & 0.0340 & $+0.0197$ \\
Two-way additive & 0.0176 & $+0.0032$ \\
Logistic fixed effects & 0.0181 & $+0.0038$ \\
\textbf{Reverse IRT} & \textbf{0.0143} & --- \\
\bottomrule
\end{tabular}
\end{table}

\subsection{Cancer Resistance Ranking}

\begin{figure}[ht]
\centering
\includegraphics[width=0.85\linewidth]{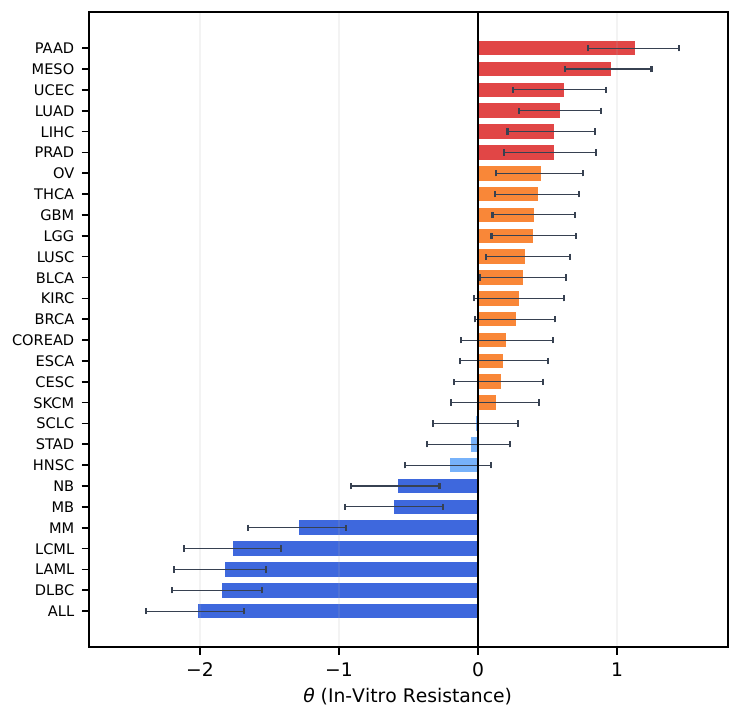}
\caption{Cancer in-vitro resistance with bootstrap 95\% CIs (200 drug-panel resamples). Red: highly resistant ($\theta > 0.5$); orange: moderately resistant; light blue: moderately sensitive; dark blue: highly sensitive. CIs crossing zero indicate uncertain classification.}
\label{fig:leaderboard}
\end{figure}

The ranking shows face-valid concordance with known clinical difficulty patterns (Figure~\ref{fig:leaderboard}). Pancreatic adenocarcinoma (PAAD) ranks most resistant, directionally consistent with clinical difficulty (13.7\% five-year relative survival; SEER). Hematological malignancies rank most sensitive, consistent with therapeutic advances in ALL (${\sim}90\%$ childhood cure rate; NCI PDQ). Nineteen of 28 cancer types have stable resistant/sensitive classifications (CIs not crossing zero); nine middle-tier cancers remain uncertain.

\subsection{External Replication}

\begin{figure}[ht]
\centering
\includegraphics[width=0.72\linewidth]{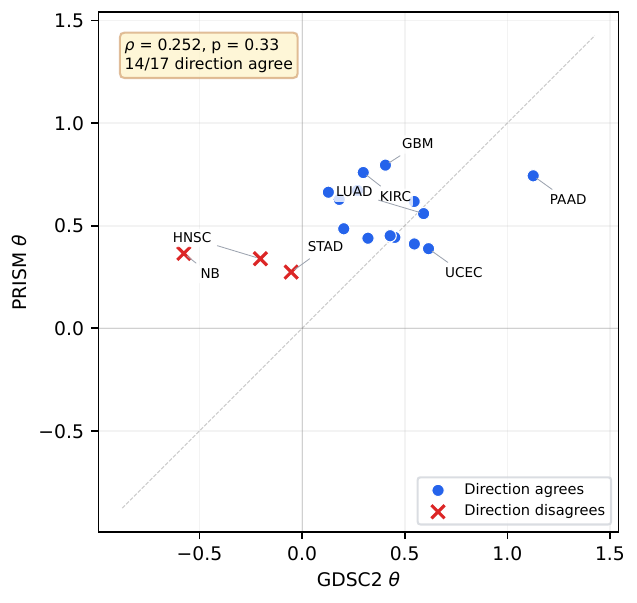}
\caption{GDSC2 vs PRISM resistance estimates. Blue: directional agreement; red crosses: disagreement. Only extremes and disagreements are labeled; full table in repository outputs. Weak rank-order correlation ($\rho = 0.25$) despite 82\% directional agreement.}
\label{fig:prism}
\end{figure}

PRISM replication yields $\rho = 0.252$ ($p = 0.33$) with 82\% directional agreement (14/17 cancers; Figure~\ref{fig:prism}). Directional agreement is strongest at the extremes. Three middle-tier cancers (STAD, HNSC, NB) show disagreement, consistent with the bootstrap uncertainty zone.

\FloatBarrier
\section{Discussion}

The primary finding is methodological: reverse IRT provides sparsity-robust ranking recovery in drug-response matrices. The Evaluation Failure Scaling Law mechanism \citep{kang2026efsl}, originally demonstrated in AI benchmark evaluation, transfers to pharmacogenomic data.

\textbf{Limitations.} (1)~Cell-line in-vitro resistance does not equal clinical resistance. (2)~Binarization at a global $\text{LN\_IC50}$ median discards continuous information. (3)~The 1PL model assumes unidimensional resistance; the LLTM \citep{fischer1973,kang2026lltm} with mutation features could decompose resistance into interpretable components. (4)~PRISM replication is directionally consistent but rank-order weak, reflecting platform differences.

\section{Conclusion}

Reverse IRT provides a sparsity-robust framework for ranking cancer types and drugs on a shared latent scale. The method outperforms averaging under all tested missingness regimes, achieves the best held-out calibration among five evaluated methods, and produces rankings with face-valid clinical concordance. The contribution is methodological: when drug-response matrices become fragmented, reverse IRT preserves ranking structure better than averaging. Cross-platform replication confirms this is a ranking methodology contribution, not a universal biological discovery.

\FloatBarrier
\section*{Data and Code Availability}

GDSC2 Release 8.5 is available from CancerRxGene:
\begin{quote}
\url{https://www.cancerrxgene.org/downloads/bulk_download}
\end{quote}
PRISM secondary dose-response and DepMap cell line metadata are available from DepMap:
\begin{quote}
\url{https://depmap.org/repurposing}
\end{quote}
All code, validation scripts, output CSVs, and metadata:
\begin{quote}
\url{https://github.com/testofschool/reverse-irt-cancer}
\end{quote}

\bibliographystyle{plainnat}

\end{document}